# GAUSSIAN KERNEL IN QUANTUM LEARNING

ARIT KUMAR BISHWAS*
AIIT, Amity University Uttar Pradesh,
Noida, India
aritkumar.official@gmail.com

ASHISH MANI
EEE, ASET, Amity University Uttar Pradesh,
Noida, India
amani@amity.edu

VASILE PALADE
Faculty of Engineering, Environment and Computing,
Coventry University, Coventry, UK
vasile.palade@coventry.ac.uk

The Gaussian kernel is a very popular kernel function used in many machine learning algorithms, especially in support vector machines (SVMs). It is more often used than polynomial kernels when learning from nonlinear datasets, and is usually employed in formulating the classical SVM for nonlinear problems. In [3], *Rebentrost et al.* discussed an elegant quantum version of a least square support vector machine using quantum polynomial kernels, which is exponentially faster than the classical counterpart. This paper demonstrates a quantum version of the Gaussian kernel and analyzes its runtime complexity using the quantum random access memory (QRAM) in the context of quantum SVM. Our analysis shows that the runtime computational complexity of the quantum Gaussian kernel is approximated to $O\ [\epsilon^{-1} dlogN]$, with error $R_d(X_{qtm})$, and even $\approx O(\epsilon^{-1} logN)$ when $d$ and the error $R_d(X_{qtm})$are small enough to be ignored, where $N$ is the dimension of the training instances, $\epsilon$ is the accuracy, $X_{qtm}$ is the dot product of the two quantum states, and $R_d(X_{qtm})$ is the Taylor remainder error term. Therefore, the run time complexity of the quantum version of the Gaussian kernel seems to be significantly faster as compared to its classical version.

## 1. Introduction

Machine learning deals with large datasets, and designs models to recognize hidden patterns in data and predict future outcomes. It is not so challenging to deal with linear datasets, which should be a straightforward process in most cases. However, when the training dataset is nonlinear, straight and simple approaches do not work well. To address nonlinear training datasets, we often use kernel functions. Applying a kernel function is a mathematical way to map a training dataset onto a higher dimensional space from a lower dimensional space. There are many popular kernel functions available, but one of the most famous is the Gaussian kernel. The Gaussian kernel is very popular in the formulation of support vector machine (SVM) for nonlinear problems. Regression analysis is also a very important field in machine learning, and, in many cases, researchers have also applied kernel tricks to deal with regression problems **[1] [2]**. Recently we have witnessed the development of some important quantum algorithms, which outperform their classical counterparts in terms of time complexity **[3] [4] [5]**. In **[3]**, authors have discussed a quantum SVM algorithm, where they used a quantum version of the polynomial kernel to handle nonlinear classification. Recently, another group of researchers discussed calculating rapidly specific nonlinear kernel functions by using both canonical and generalized coherent states **[6]**.

This paper proposes a formulation of a quantum version of the Gaussian kernel. A Gaussian kernel is the normalized polynomial kernel of infinite degree. We followed the same notion in quantum setting and used normalized quantum polynomial kernels to formulate the quantum Gaussian kernel. The quantum version of the Gaussian kernel exhibits $O(\epsilon^{-1} dlogN)$ runtime complexity with $N-dimensional$ instances, an accuracy $\epsilon$ and with an error $\left|R_d(X_{qtm})\right|$, where $R_d$ is the remainder term of a specific Taylor series. Our proposed quantum Gaussian kernel shows significantly better performance than its classical counterpart does in terms of time complexity. The proposed quantum Gaussian kernel formulation can be a tool in many quantum based applications in machine learning, broadly in quantum classification (for example, quantum SVM **[3]**) and quantum clustering.

## 2. Gaussian Kernel in the Classical Approach

We begin our discussion by presenting the Gaussian kernel formulation in the classical approach. Consider a supervised learning problem with a set of training examples, $\{X_i, X_j\}$, which consists of $M$ $N-$dimensional inputs $X_i$, and associated outputs, $Y_i$. We define a polynomial kernel as:

$$K_{polyK}(X_i, X_j) = \left(< X_i, \quad X_j >\right)^d \approx (A(X_i^T X_j) + B)^d; A > 0, B \geq 0 \qquad (1)$$

Where $A$ is the scaling factor which scales the $d-$degree polynomial kernel by $\sqrt{A}$, and $B$ is a free parameter, which trades off the influence between the lower-order terms and higher-order terms in the polynomial. We define a function $K_{GK}(X_i, X_j)$, which is the sum of infinite polynomial kernel series, as follows:

$$K_{GK}(X_i, X_j) = 1 + \frac{(X_i^T X_j)}{1!} + \frac{(X_i^T X_j)^2}{2!} + \frac{(X_i^T X_j)^3}{3!} + \cdots to\ \infty = \sum_{l=0}^{\infty} \frac{(X_i^T X_j)^l}{l!} \approx e^{< X_i, X_j>} \qquad (2)$$

We consider, $A = 1, B = 0$, for simplifying the calculation. In addition, here we are mostly interested in the dot product evaluation and can avoid the constants, $A$ and $B$, in the calculation.

We elaborate the Equation.2 as follows:

$$\begin{aligned} K_{GK}(X_i, X_j) &= e^{\left(-\frac{|X_i - X_j|^2}{2}\right)} = e^{\left[-\frac{1}{2}<(X_i - X_j),(X_i - X_j)>\right]} = e^{\left[-\frac{1}{2}(<X_i, X_i - X_j> - <X_j, X_i - X_j>)\right]} \\ &= e^{\left[-\frac{1}{2}(<X_i, X_i> - <X_i, X_j> - <X_j, X_i> - <X_j, X_j>)\right]} = e^{\left[-\frac{1}{2}\left(|X_i|^2 - |X_j|^2 - 2< X_i, X_j>\right)\right]} \\ &= e^{\left[-\frac{1}{2}\left(|X_i|^2 - |X_j|^2\right)\right]} e^{\left[-\frac{1}{2}(-2< X_i, X_j>)\right]} = C\left[e^{\left\{-\frac{1}{2}(-2< X_i, X_j>)\right\}}\right] \\ &= C\left[e^{< X_i, X_j>}\right] \approx \sum_{l=0}^{\infty} \frac{(X_i^T X_j)^l}{l!} = e^{(X_i^T X_j)} \end{aligned} \qquad (3)$$

where, $C = e^{\left[-\frac{1}{2}\left(|X_i|^2 - |X_j|^2\right)\right]}$ is a constant.

In the equation (3), we induce a parameter $\sigma$ in such a way that it scales the input vectors by a factor of $\frac{1}{\sigma}$, we get

$$K_{GK}(X_i, X_j) = e^{\left(-\frac{|X_i - X_j|^2}{2\sigma^2}\right)} \qquad (4)$$

The equation (4) is the formulation of the Gaussian kernel with parameter $\sigma$.

We now analyze the approximated runtime complexity of calculating the Gaussian kernel.

$$runtime\ complexity\ of\ K_{GK}(X_i, X_j) = runtime\ complexity\ of\ (1) + runtime\ complexity\ of\ \left(\frac{(x_i^T x_j)^1}{2!}\right) + runtime\ complexity\ of\ \left(\frac{(x_i^T x_j)^2}{2!}\right) + \cdots to\ \infty \quad (5)$$

As we know, in a dot product of two $N - dimensional$ vectors, we perform $N$ multiplications and $(N - 1)$ additions. Here, multiplication and addition are constant-time operations, therefore the time-complexity of the dot product is $O(N) + O(N) = O(N)$. So, the runtime complexity of calculating the Gaussian kernel (referring to Equations 3, 4 and 5) will be:

$$runtime\ complexity\ of\ K_{GK}(X_i, X_j) = runtime\ complexity\ of\ \left[1 + \frac{(x_i^T x_j)}{1!} + \frac{(x_i^T x_j)^2}{2!} + \frac{(x_i^T x_j)^3}{3!} + \cdots to\ \infty\right] \quad (6)$$

Here, $1 + \frac{(x_i^T x_j)}{1!} + \frac{(x_i^T x_j)^2}{2!} + \frac{(x_i^T x_j)^3}{3!} + \cdots to\ \infty$ is an infinite Taylor series, and we cut down the infinite series by inducing a remainder term, $R_d(X_i^T X_j)$, which helps in approximating the infinite Taylor series into a $d^{th} - degree$ finite series with error $R_d(X_i^T X_j)$. This results in to:

$$runtime\ complexity\ of\ K_{GK}(X_i, X_j) = runtime\ complexity\ of\ 1 + \frac{(x_i^T x_j)}{1!} + \frac{(x_i^T x_j)^2}{2!} + \frac{(x_i^T x_j)^3}{3!} + \cdots + \frac{(x_i^T x_j)^d}{d!} + R_d(X_i^T X_j) \cong O\left[\left(1 + \frac{(N)^1}{1!} + \frac{(N)^2}{2!} + \frac{(N)^3}{3!} + \cdots + \frac{(N)^d}{d!}\right) + R_d(X_i^T X_j)\right] \approx O[Poly(N)]\ with\ error\ R_d(X_i^T X_j) \quad (7)$$

where $\left|R_d(X_i^T X_j)\right| = \left|\frac{e^{\xi}}{(d+1)!}(X_i^T X_j)^{d+1}\right| < 10^{-q}, 0 < \xi < N, -1 \le X_i^T X_j \le 1, 0 \le d$.

In equation (7), the series approximately converges up to $q$ correct decimal places.

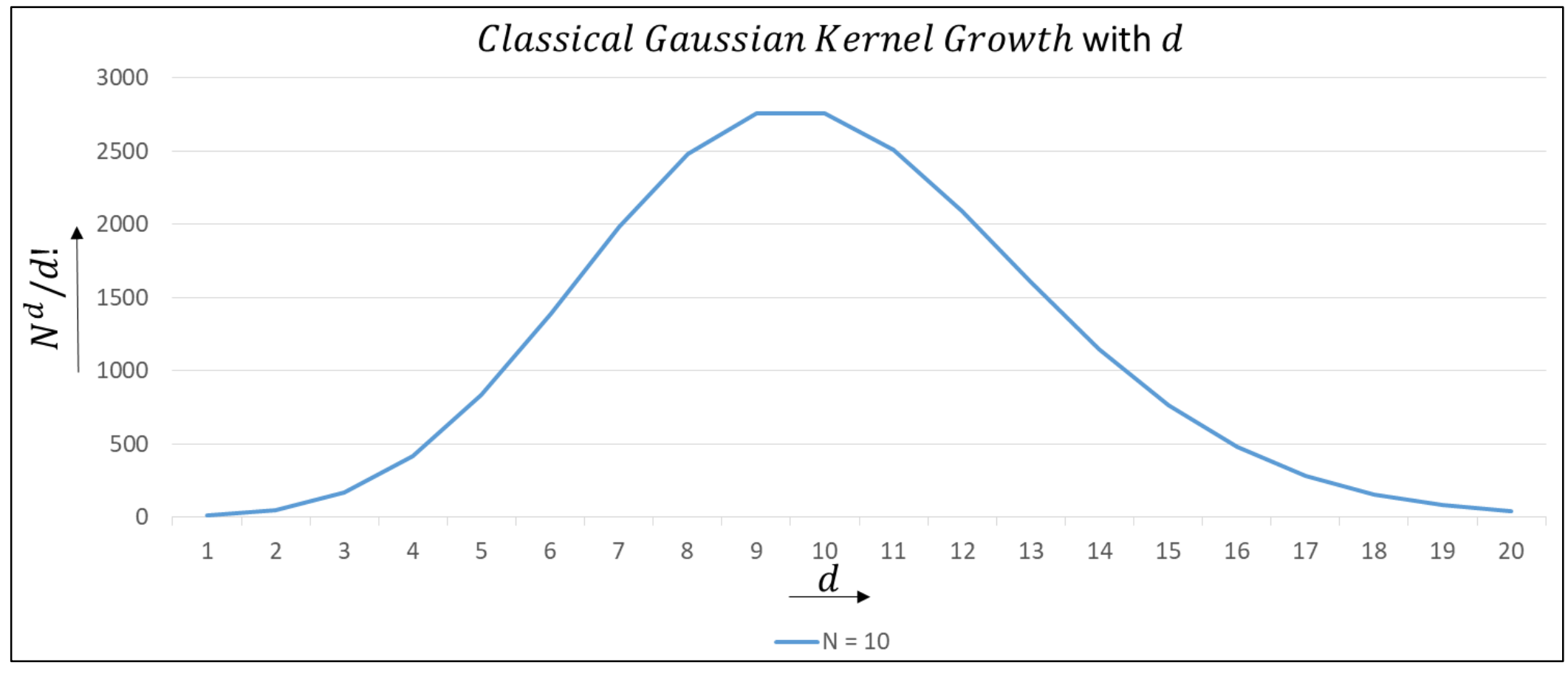

Fig. 1. Classical Gaussian Kernel growth with $d$ for $N = 10$

Fig.1. shows that the relationship between $\left(\frac{N^d}{d!}\right)$ and $d$. Here, $Poly(N)$ is a $d-degree$ polynomial and associated with $O(dN)$ runtime complexity (*using Horner's Rule* **[7]**). Therefore, we can approximate the upper bound of the overall runtime complexity of the classical Gaussian kernel as:

$$\approx O\ (dN),\ with\ error\ R_d(X_i^T X_j) \tag{8}$$

We now discuss the error part, the remainder $\left|R_d(X_i^T X_j)\right| = \left|\frac{e^{\xi}}{(d+1)!}\left(X_i^T X_j\right)^{d+1}\right|$. With $R_d\left(X_i^T X_j\right) < 10^{-q}$, we ensure that the error in the approximation is no more than $10^{-q}$. The approximated kernel calculation is precise up to $q^{th}$ decimal places. Assuming $-1 < X_i^T X_j \le 1$, we define:

$$e^{X_i^T X_j} = 1 + X_i^T X_j + \frac{e^{\xi}}{2}\left(X_i^T X_j\right)^2 < 1 + X_i^T X_j + \frac{e^{X}}{2}\left(X_i^T X_j\right)^2,\ 0 < X_i^T X_j \le 1, e^{\xi} < e^{X_i^T X_j}, 0 < \xi < X_i^T X_j \tag{9}$$

$$e^{X_i^T X_j} - \frac{e^{X_i^T X_j}}{2}\left(X_i^T X_j\right)^2 \le 1 + X_i^T X_j$$

$$\Rightarrow\ e^{X_i^T X_j}\left(1 - \frac{1}{2}\left(X_i^T X_j\right)^2\right) \le 1 + X_i^T X_j$$

$$\Rightarrow\ e^{X_i^T X_j} \le \frac{\left(1+\left(X_i^T X_j\right)\right)}{\left(1-\frac{1}{2}\left(X_i^T X_j\right)^2\right)} = \frac{2\left(1+\left(X_i^T X_j\right)\right)}{\left(2-\left(X_i^T X_j\right)^2\right)} \le 4,\ 0 \le X_i^T X_j \le 1 \tag{10}$$

By using the above inequality (10) and $e^{\xi} < e^{X_i^T X_j}, \forall\ X_i^T X_j \in [-1,0]$, we get

$$R_d\left(X_i^T X_j\right) = \frac{e^{\xi}}{(d+1)!}\left(X_i^T X_j\right)^{d+1} \le \frac{e^{X}}{(d+1)!}\left(X_i^T X_j\right)^{d+1} \le \frac{4}{(d+1)!}\left(X_i^T X_j\right)^{d+1} \le \frac{4}{(d+1)!},\ -1 < X_i^T X_j \le 1 \tag{11}$$

The required precision is indeed touched when

$$\frac{4}{(d+1)!} < 10^{-q} \Leftrightarrow 4\ (10^{q}) < (d+1)! \tag{12}$$

For example: with $q = 5$,

$$4\ (10^{5}) < (d+1)! \Rightarrow 9 \le d.$$

In this example, the approximation provides a calculation correct up to five decimal places with $9 \le d$. In a similar way, the approximation provides a calculation correct up to four decimal places with $8 \le d$.

## 3. Quantum Random Access Memory (QRAM)

Before discussing the quantum Gaussian kernel, it is very important to understand the concept of QRAM **[8-14],** as the proposed quantum Gaussian kernel works with QRAM. A QRAM is the quantum version of classical RAM. It contains the address and output registers, which are composed of qubits. The QRAM allows accessing the data in a quantum parallel way and performing memory access in coherent quantum superposition **[15]**. The address register $R_{ADR}$ holds a superposition of addresses. The QRAM returns a superposition of data in a data register $DR$ as output, which is correlated with the address register $R_{ADR}$,

$$\sum_j \psi_j |j\rangle_{R_{ADR}} \rightarrow \sum_j \psi_j |j\rangle_{ADR} \, |D_j\rangle_{DR} \tag{13}$$

where, $R_{ADR}$ contains a superposition of addresses $\sum_j \psi_j |j\rangle_{R_{ADR}}$, and $D_j$ is the $j^{th}$ memory cell content. It takes $O(log\ N)$ steps to reconstruct any quantum state from QRAM, where $N$ is the dimension of the complex vector.

## 4. Quantum Gaussian Kernel

In Section 2, we have discussed how a classical Gaussian kernel can be formulated using polynomial kernels. In a quantum setting, we can formulate the Gaussian kernel using quantum polynomial kernels too. Consider $K_{qGK}\left(|X_i\rangle, |X_j\rangle\right)$ is a Gaussian kernel, where $|X_i\rangle$ and $|X_j\rangle$ are the $N - dimensional$ training input vectors in a quantum form. We now formulate the quantum Gaussian kernel using quantum polynomial kernels as follows:

$$K_{qGK}\left(|X_i\rangle, |X_j\rangle\right) = 1 + \frac{\langle X_i|X_j\rangle}{1!} + \frac{\langle X_i|X_j\rangle^2}{2!} + \frac{\langle X_i|X_j\rangle^3}{3!} + \cdots to \infty = \tag{14}$$

$$= \sum_{l=0}^{\infty} \frac{\langle X_i|X_j\rangle^l}{l!} = e^{\langle X_i|X_j\rangle} \tag{15}$$

Where, $\langle X_i|X_j\rangle^d$ is a quantum polynomial kernel with degree $d$ (for $d = 1$, it will be a linear kernel).

After normalizing the above equation (15), we get:

$$K_{qGK}(|\vec{X}_i\rangle, |\vec{X}_j\rangle) = \frac{e^{\langle X_i|X_j\rangle}}{\sqrt{e^{\langle X_i|X_i\rangle}}\sqrt{e^{\langle X_j|X_j\rangle}}}$$

$$= e^{\langle X_i|X_j\rangle - \frac{1}{2}\langle X_i|X_i\rangle - \frac{1}{2}\langle X_j|X_j\rangle}$$

$$= e^{\langle X_i|X_j\rangle - \frac{1}{2}[\langle X_i|X_i\rangle + \langle X_j|X_j\rangle]}$$

$$= e^{\frac{2\langle X_i|X_j\rangle - [\langle X_i|X_i\rangle + \langle X_j|X_j\rangle]}{2}}$$

$$= e^{-\frac{1}{2}[\langle X_i|X_i\rangle + \langle X_j|X_j\rangle - 2\langle X_i|X_j\rangle]}, \tag{16}$$

where $|\vec{X}_i\rangle$ and $|\vec{X}_j\rangle$ are normalized vectors. In the similar notion of classical formulation in section 2, the parameter $\sigma$ can be used in quantum setting too, and the number of contours in the higher dimensional feature space can be controlled with this $\sigma$ parameter,

$$K_{qGK}\left(|\vec{X}_i\rangle, |\vec{X}_j\rangle\right) = e^{-\frac{1}{2\sigma^2}[\langle X_i|X_i\rangle + \langle X_j|X_j\rangle - 2\langle X_i|X_j\rangle]}$$

$$= e^{-\frac{1}{2\sigma^2}[\langle X_i|X_i\rangle + \langle X_j|X_j\rangle]} . e^{-\frac{1}{2\sigma^2}[-2\langle X_i|X_j\rangle]}$$

$= \text{C}.e^{-\frac{1}{2\sigma^2}\left[-2\langle X_i|X_j\rangle\right]} \approx e^{\langle X_i|X_j\rangle},$ (17)

where $\text{C} = e^{-\frac{1}{2\sigma^2}\left[\langle X_i|X_i\rangle + \langle X_j|X_j\rangle\right]}$ is a constant. In the above equations (15) and (17), the main idea is to evaluate a dot product of the two training inputs quantum mechanically. Once we evaluate the dot products quantum mechanically, we can calculate the $d$ −degree polynomial kernels, and hence the Gaussian kernel.

We now discuss a quantum mechanical process for dot product evaluation of two normalized training inputs, $|\vec{X}_i\rangle$ *and* $|\vec{X}_j\rangle$ (in quantum form), for the linear kernel. For evaluating a dot product of $|\vec{X}_i\rangle$ *and* $|\vec{X}_j\rangle$, first of all, we generate two quantum states $|\psi\rangle$ *and* $|\phi\rangle$ with an ancilla variable **[16] [17]**. We then estimate the sum of the squared norms of the two training inputs, say parameter $Z = \|\vec{X}_i\|^2 + \|\vec{X}_j\|^2$. At the end, we do a swap test to perform a projective measurement on the ancilla alone.

Therefore, initially, we construct a quantum state $|\psi\rangle$ by querying the QRAM:

$$|\psi\rangle = \frac{1}{\sqrt{2}}\left(|0\rangle|\vec{X}_i\rangle + |1\rangle|\vec{X}_j\rangle\right) \quad (18)$$

Let us consider another quantum state

$$|\xi\rangle = \left(\frac{1}{\sqrt{2}}(|0\rangle - |1\rangle) \otimes |0\rangle\right) \quad (19)$$

We apply a unitary transformation **[17]** $e^{-iHt}$ to the state $|\xi\rangle$ ., where $H = \left(\left(\|\vec{X}_i\||0\rangle\langle 0| + \|\vec{X}_j\||1\rangle\langle 1|\right) \otimes \sigma_x\right)$ is a Hamiltonian, and $\sigma_x = \begin{bmatrix} 0 & 1 \\ 1 & 0 \end{bmatrix}$ is a Pauli spin matrix. This results in the following state:

$$\left[\frac{1}{\sqrt{2}}\left(cos\left(\|\vec{X}_i\|t\right)|0\rangle - cos\left(\|\vec{X}_j\|t\right)|1\rangle\right) \otimes |0\rangle\right] - \left[\frac{i}{\sqrt{2}}\left(sin\left(\|\vec{X}_i\|t\right)|0\rangle - sin\left(\|\vec{X}_j\|\right)|1\rangle\right) \otimes |1\rangle\right] \quad (20)$$

Now, we measure the ancilla bit with an appropriate choice of $t, where\ \|\vec{X}_i\|t, \|\vec{X}_j\|t \ll 1$, which results in the state:

$$|\phi\rangle = \frac{1}{\sqrt{\left(\|\vec{X}_i\|^2 + \|\vec{X}_j\|^2\right)}}\left(\|\vec{X}_i\||0\rangle - \|\vec{X}_j\||1\rangle\right) \quad (21)$$

with probability $\frac{1}{2}\left[\left(\|\vec{X}_i\|^2 + \|\vec{X}_j\|^2\right)t\right]^2$ **[17]**.

This allows the estimation of the sum of the squared norms of $|\vec{X}_i\rangle$ *and* $|\vec{X}_j\rangle$. By using quantum counting **[15],** we can estimate $\left(\|\vec{X}_i\|^2 + \|\vec{X}_j\|^2\right)$ and create the quantum state $|\phi\rangle$ with accuracy $\epsilon$, and therefore the complexity will be $O(\epsilon^{-1})$. We now perform a swap test with states $|\psi\rangle$ *and* $|\phi\rangle$ using an ancilla alone. If $|\psi\rangle$ $and$ $|\phi\rangle$ are equal, then the measurement give us a zero. Thus, the overall complexity to evaluate a single dot product of the training instances considering the QRAM access **[8]**, estimating $\left(\|\vec{X}_i\|^2 + \|\vec{X}_j\|^2\right)$ and constructing the quantum state $|\phi\rangle$, is

$$O(\epsilon^{-1} logN) \tag{22}$$

We now generalize the context for non-linear polynomial kernel function. We consider $d$ copies of $|\vec{X}_i\rangle$ $and$ $|\vec{X}_j\rangle$. Each instance maps into the $d$-times tensor product, and the polynomial kernel is formulated by mapping the original instances of the linear space to the $d$-dimensional linear hyperspace. We simply map each vector $|\vec{X}_i\rangle$ into the $d$ −times tensor product $|\vec{X}_i\rangle \otimes |\vec{X}_i\rangle \otimes \ldots d\ times = |\vec{X}_i\rangle^{\otimes d}$. Therefore, a $d-$ degree polynomial kernel can be constructed as $\langle\vec{X}_i|\vec{X}_j\rangle^d$ **[17]**. Thus, the overall runtime complexity of a $d$ −degree polynomial kernel is:

$$O(\epsilon^{-1} dlogN) \tag{23}$$

We now evaluate the complexity of the quantum Gaussian kernel $K_{qGK}\left(|\vec{X}_i\rangle, |\vec{X}_j\rangle\right)$. We have calculated the complexity of each quantum polynomial kernel in the infinite series, and solve the summation of the series for extracting the overall run time complexity. Referring to the equations (15-17),

$$runtime\ complexity\ of\ K_{qGK}\left(|\vec{X}_i\rangle, |\vec{X}_j\rangle\right) = runtime\ complexity\ of\ \left(\frac{\langle\vec{X}_i|\vec{X}_j\rangle}{1!}\right) + runtime\ complexity\ of\ \left(\frac{\langle\vec{X}_i|\vec{X}_j\rangle^2}{2!}\right) + \cdots to\ \infty \tag{24}$$

Here, $\langle\vec{X}_i|\vec{X}_j\rangle^d$ is the $d-degree$ quantum polynomial kernel.

$$runtime\ complexity\ of\ \left(\frac{\langle\vec{X}_i|\vec{X}_j\rangle}{1!}\right) + runtime\ complexity\ of\ \left(\frac{\langle\vec{X}_i|\vec{X}_j\rangle^2}{2!}\right) + \cdots to\ \infty = runtime\ complexity\ of\ \left[\left(\frac{\langle\vec{X}_i|\vec{X}_j\rangle}{1!}\right) + \left(\frac{\langle\vec{X}_i|\vec{X}_j\rangle^2}{2!}\right) + \cdots + \left(\frac{\langle\vec{X}_i|\vec{X}_j\rangle^d}{d!}\right)\right] with\ error\ R_d\left(X_{qtm}\right), \tag{25}$$

where $X_{qtm} = \langle\vec{X}_i|\vec{X}_j\rangle$.

By using equation (23), we get

$$runtime\ complexity\ of\ \left[\left(\frac{\langle\vec{X}_i|\vec{X}_j\rangle}{1!}\right) + \left(\frac{\langle\vec{X}_i|\vec{X}_j\rangle^2}{2!}\right) + \cdots + \left(\frac{\langle\vec{X}_i|\vec{X}_j\rangle^d}{d!}\right)\right] with\ error\ R_d\left(X_{qtm}\right) = O\left[\frac{(\epsilon^{-1}logN)}{1!} + \frac{(\epsilon^{-1}2logN)}{2!} + \frac{(\epsilon^{-1}3logN)}{3!} + \cdots + \frac{(\epsilon^{-1}dlogN)}{d!}\right] with\ error\ R_d\left(X_{qtm}\right) \tag{26}$$

Equation (26) is actually an infinite Taylor series; here we cut down the infinite series by inducing a remainder term $R_d(X_{qtm})$, which helps in approximating the infinite Taylor series into a finite series of $d$ terms, with error $R_d(X_{qtm})$.

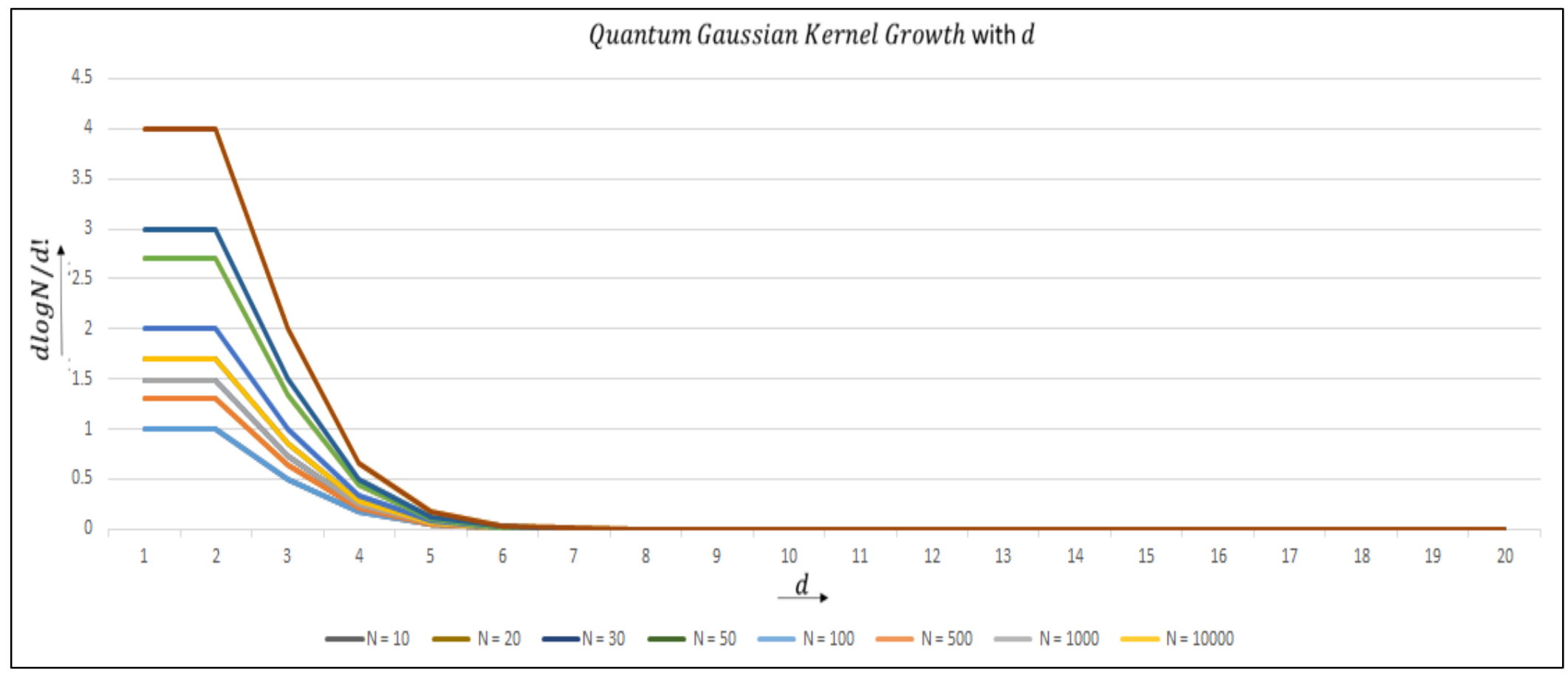

Fig. 2. Quantum Gaussian Kernel growth with $d$ for different values of $N$. We did not include the $\epsilon$ in the calculation for simplicity, although it does not change the context much here.

The plot $\left(\frac{dlogN}{d!}\right) vs.d$ in **Fig.2** shows that after few terms (~7 in the above Figure 2), the later terms in the series almost vanish (not completely although). Therefore, the upper bound of the equation (26) can be approximately restricted to (by using equation (26)):

$$O[\epsilon^{-1}dlogN] \text{ } with \text{ } error \text{ } R_d\left(X_{qtm}\right), \tag{27}$$

where $\left|R_d(X_{qtm})\right| < 10^{-q}$, and here the definition of $R_d(X_{qtm})$ is similar to the discussion done in Section 2. In the equation (27), the series approximately converges up to $q$ correct decimal places. Even $O[\epsilon^{-1}dlogN] \text{ } with \text{ } error \text{ } R_d\left(X_{qtm}\right) \approx O(\epsilon^{-1}logN)$, if $d$ is small enough and we can ignore the error $R_d(X_{qtm})$ for simplicity. Therefore, the overall complexity of the proposed quantum Gaussian Kernel $K_{GK}(|\vec{X}_i\rangle, |\vec{X}_j\rangle)$ is approximated to $O(\epsilon^{-1}dlogN)$, with error $R_d\left(X_{qtm}\right)$. Compared to the classical counterpart, as explained in Section 2, the proposed quantum Gaussian kernel is significantly faster in terms of computational complexity.

## 5. Applications

This section discusses some application of the proposed quantum Gaussian kernel. In **[3] [18] [19]**, authors discussed a quantum version of SVMs. The Quantum SVM potentially exhibits an exponential faster runtime as compared to the classical counterpart. In **[3] [18] [19]**, authors have also used the quantum linear and quantum polynomial kernels to formulate the quantum least square SVM, where a kernel function is vital. The Gaussian kernel defines a function (exponential in nature), which possesses a much larger function space as compared to the polynomial kernel. As we enhance the order of the polynomial function, the function space is increased. Therefore, for an $m^{th}$ order polynomial kernel, all the derivatives of higher order ($> m + 1$) become zero.

However, with the Gaussian kernel, the function space is infinite, being an infinite order polynomial kernel. In addition, models with a polynomial kernel are parametric in nature, where the complexity of the models are fixed and bounded (even if the amount of data is unbounded), and become saturated after a certain period. Models with Gaussian kernels are non-parametric in nature, where the complexity of the model can grow with the growth of input data. Therefore, technically, we can conclude that Gaussian kernels can handle more complex nonlinear data structure than linear/polynomial kernels.

In the quantum SVM formulation, we achieve an exponential speed up in the following two ways: a) by calculating the kernel matrix in a quantum way **[3][20]**, and b) with the speed up gain in the number of training examples by obtaining the optimum of the least squares dual formulation . From the context of interest, we discuss only the former here. With a quantum polynomial kernel of order $d$, the overall complexity of the quantum SVM is $O\ (M^3 + M^2 d\epsilon^{-1} logN)$. At this point, we see a speed up gain in the overall complexity, due to the fast kernel matrix calculation in the quantum way (as discussed in Section 4) **[20]**; where $M$ is the number of training examples and $N$ is the dimension of the feature space.

In more details, when we apply the proposed quantum Gaussian kernel instead of the quantum linear/polynomial kernels **[3] [18] [19]** in quantum SVM formulation, the runtime complexity of the quantum SVM with kernel matrix calculation becomes:

$$O\ (M^3 + M^2 \epsilon^{-1} dlogN)\ with\ error\ R_m\left(X_{qtm}\right), \tag{28}$$

and $O\ (M^3 + M^2 \epsilon^{-1} logN)$, when we ignore the error $R_d(X_{qtm})$ for simplicity and $d$ is considered small.

## 6. Conclusions

This paper analyzed the complexity of calculating the quantum Gaussian kernel and it was shown that its runtime complexity is approximated to $O(\epsilon^{-1} dlogN)$, which is significantly faster than the classical counterpart. The analysis also suggests some reasons for the good performance of Gaussian kernel over the polynomial kernel, because of employing an infinite dimensional polynomial kernel within it. The Gaussian kernel is a very popular choice and has been used in a broad range of applications. Therefore, it can be a better choice to tackle the learning of non-linear datasets in quantum SVM, as compared to using quantum polynomial kernels.